\documentclass[journal]{IEEEtran}

\usepackage{graphicx,epsfig}
\usepackage{amsmath,mathrsfs,xcolor} 
\usepackage{algorithm}
\usepackage[noend]{algpseudocode}
\makeatletter
\def\BState{\State\hskip-\ALG@thistlm}
\makeatother
\usepackage{amssymb} 
\usepackage{amsthm}
\usepackage{amsfonts}
\usepackage{mathtools}
\usepackage{multirow}
\usepackage{mathtools} 
\usepackage{cite}
\usepackage{epstopdf}
\usepackage{ifthen}
\usepackage{bbm}
\usepackage{setspace}

\newcommand{\R}{ \mathbb{R}}

\newcommand{\bz}{{\boldsymbol z}}
\newcommand{\bx}{{\boldsymbol x}}
\newcommand{\by}{{\boldsymbol y}}

\newcommand{\bn}{{\boldsymbol p}}

\newcommand{\cnn}{C\!N\!N}

\def\Spc#1{{\mathcal{#1}}}  
\def\M#1{{\bf{#1}}}  

\DeclareMathOperator*{\argmin}{arg\,min}

\providecommand{\norm}[1]{\left\rVert#1\right\rVert}

\def\R{{\mathbb{R}}}

\def\dimx{{\mathbb{R}^N}}
\def\dimH{{\mathbb{R}^{M\times N}}}
\def\dimy{{\mathbb{R}^{M}}}

\def\btheta{{\boldsymbol{\theta}}}
\def\bx{{\rm \bf x}}
\def\by{{\rm \bf y}}
\def\bz{{\rm \bf z}}
\def\br{{\rm \bf r}}
\def\bn{{\rm \bf n}}
\def\bH{{\rm\bf H}}
\def\HTH{{\bH^{\T}\bH}}
\def\HT{{\bH^{\T}}}
\def\bI{{\rm\bf I}}

\def\cS{{\mathcal{S}}}
\def\cT{{\mathcal{T}}}
\def\cB{{\mathcal{B}}}
\def\cC{{\mathcal{C}}}

\def\half{{\frac{1}{2}}}

\def\G{G}
\def\F{F}

\newcommand{\Frac}[2]{{#1}/{#2} }
\newcommand{\ip}[2]{\left\langle {#1}\, , \,{#2}\right\rangle}

\makeatletter
\DeclareRobustCommand\onedot{\futurelet\@let@token\@onedot}
\def\@onedot{\ifx\@let@token.\else.\null\fi\xspace}
\def\eg{\emph{e.g}\onedot} 
\def\ie{\emph{i.e}\onedot} 
 
\def\etc{\emph{etc}\onedot} 
\def\wrt{w.r.t\onedot} 

\makeatother
\newtheorem{thm}{Theorem}
\newtheorem{prop}{Proposition}

\DeclareMathOperator{\id}{Id}

\DeclareMathOperator{\T}{T}

\usepackage{ dsfont,color }
\usepackage{slashbox}
\usepackage{array,multirow}
\usepackage{diagbox}
\usepackage{graphicx,epsfig}

\usepackage{framed}

\usepackage{float}
\newfloat{algorithm}{t}{lop}
\usepackage{amsfonts,amsmath,amssymb,graphicx,color,epstopdf,amsthm}
\usepackage{cite}
\usepackage{float} 
\usepackage{xspace}
\usepackage{url}
\usepackage{booktabs} 
\usepackage{tikz}
\usetikzlibrary{positioning,calc}
\usepackage[caption=false]{subfig} 

\begin{document}
\IEEEoverridecommandlockouts



\title{CNN-Based Projected Gradient Descent for Consistent Image Reconstruction}

\author{Harshit~Gupta,~Kyong~Hwan~Jin,~Ha Q. Nguyen
,~Michael~T.~McCann,
        and~Michael~Unser~\IEEEmembership{,~IEEE Fellow}
        \thanks{Authors are with the Biomedical Imaging Group, EPFL, Lausanne, Switzer-
land. This project is funded by H2020-ERC, grant agreement No. 692726 - GlobalBioIm. 

This manuscript was submitted to ``IEEE Transactions on Medical Imaging'', for the special edition, ``Machine Learning for Image Reconstruction'', on 30 Aug. 2017.}}



\maketitle


\begin{abstract}
We present a new method for image reconstruction which replaces the projector in a projected gradient descent (PGD) with a convolutional neural network (CNN).
CNNs trained as high-dimensional (image-to-image) regressors have recently been used to efficiently solve inverse problems in imaging.
However, these approaches lack a feedback mechanism to enforce that the reconstructed image is consistent with the measurements.
This is crucial for inverse problems, and more so in biomedical imaging, where the reconstructions are used for diagnosis.

In our scheme, the gradient descent enforces measurement consistency, while the CNN recursively projects the solution closer to the space of desired reconstruction images.
We provide a formal framework to ensure that the classical PGD converges to a local minimizer of a non-convex constrained least-squares problem.
When the projector is replaced with a CNN, we propose a relaxed PGD, which always converges.
Finally, we propose a simple scheme to train a CNN to act like a projector.

Our experiments on sparse view Computed Tomography (CT) reconstruction for both noiseless and noisy measurements show an improvement over the total-variation (TV) method and a recent CNN-based technique. 
\end{abstract}


\section{Introduction }
While medical imaging is a fairly mature area, there is recent evidence that it may still be possible to reduce the radiation dose and/or speedup the acquisition process without compromising image quality \cite{Lustig2007}.
This can be accomplished with the help of sophisticated reconstruction algorithms that incorporate some prior knowledge (\eg, sparsity) on the class of 
underlying images.
The reconstruction task is usually formulated as an inverse problem, where the image-formation physics is modeled by an operator $\M H:\R^N \to \R^M $ (called the {\emph
 {forward model}}).
The measurement equation is $\by=\bH \bx+\bn \in \dimy$, where $\bx$ is the space-domain image that we are interested in recovering and $\bn \in \dimy$ is the noise intrinsic to the acquisition process.

In the case of {\emph
 {extreme imaging}}, the number and the quality of the measurements are both reduced as much as possible, \eg, in order to decrease either the scanning time in MRI or the radiation dose in CT.
Moreover, the measurements are typically very noisy due to short integration times, which calls for some form of denoising.
Indeed, there may be significantly fewer measurements than the number of unknowns ($M << N$). 
This gives rise to an ill-posed problem in the sense that there may be an infinity of consistent images that map to the same measurements $\M y$. Thus, one challenge of the reconstruction algorithm is essentially to select the ``best'' solution among a multitude of potential candidates.

The available reconstruction algorithms can be broadly arranged in three categories (or generations), which represent the continued efforts of the research community to address the aforementioned challenges.

\subsubsection{Classical Algorithms} Here, the reconstruction is performed directly by applying a suitable linear operator. This may be the backprojection (BP) $\HT \by$ or the filtered backprojection (FBP) 
 $\M F\HT\by$, where $\M F: \dimx \to \dimx $ is a regularized version of $(\HT\bH)^{-1}$.\footnote{Or, $\M F$ can be a regularized version of $(\bH\HT)^{-1}$ and be applied as $\HT\M F$, as is often the case in CT.} 
FBP-type algorithms are fast; they provide excellent results when the number of measurements are sufficient and the noise is small \cite{Pan2009}.
However, they are not suitable for extreme imaging scenarios 
because they introduce artifacts intimately connected to the inversion step.

\subsubsection{Iterative Algorithms} These algorithms avoid the shortcomings of the classical ones by solving
\begin{equation}
\M x^*= \argmin_{\M x} (E(\M H \M x, \M y)+\lambda R(\M x)),\label{iterative}
\end{equation}
where $E:\dimy\times \dimy\to \R^+$ is a data-fidelity term that favors solutions that are consistent with the measurements, $R: \dimx \to \R^+$ is a suitable regularizer that encodes the prior knowledge about the image $\M x$ to be reconstructed, and $\lambda \in \R^+$ is a tradeoff parameter. The quantity $\by^\ast=\M H\M x^\ast$, where 
$\bx^\ast$ is the solution of \eqref{iterative}, can be interpreted as the denoised version of $\M y$.
Under the assumption that the functionals $E$ and $R$ are convex, one can show that the solution of \eqref{iterative} also satisfies
\begin{equation}
\bx^\ast=\arg \min_{\M x \in \dimx} R(\M x) \quad \text{s.t.} \quad \M H \M x=\M y_0.
\end{equation}
Therefore,
among all potential solutions that admit the denoised measurement $\by^\ast$, the algorithm picks the one with the least $R$. This shows that the quality of the reconstruction depends heavily on the prior encoder $R$. Generally, these priors are either handpicked (\eg, total variation (TV) or the $\ell_1$-norm of the wavelet coefficients of the image \cite{Bouman1993,Charbonnier1997,Lustig2007, Candes2007,Ramani2011}) or learned through a dictionary \cite{Elad2006,Candes2011,Ravishankar2017}. However, in either case, they are restricted to well-behaved functionals that can be minimized via a convex routine \cite{Figueiredo2003,daubechies2004iterative,beck2009fast,boyd2011distributed}.
This limits the type of prior knowledge that can be injected into the algorithm.

An interesting variant within the class of iterative algorithms is ``plug-and-play ADMM''\cite{venkatakrishnan2013plug}. We recall that ADMM is an iterative optimization technique \cite{boyd2011distributed}, which repeatedly alternates between:  (i) a linear solver that reinforces consistency \wrt measurements; (ii) a nonlinear operation that re-injects the prior. Interestingly, the effect of (ii) is akin to denoising. This perspective has resulted in a scheme where an off-the-shelf denoiser is plugged into the later step \cite{venkatakrishnan2013plug,chan2017plug,sreehari2016plug,chang2017,romano2016little}.
This scheme, therefore, is more general than the optimization framework \eqref{iterative} but still lacks theoretical justifications. In fact, there is no good understanding yet of the connection between the use of a given denoiser and the regularization it imposes.

\subsubsection{Learning-based Algorithms} Recently, there has been a surge in using deep learning to solve inverse problems in imaging~\cite{jin2017deep,han2017deep,antholzer2017deep,wang2016accelerating,ReviewMike}, establishing new state-of-the-art results for tasks like sparse-view CT reconstruction~\cite{jin2017deep}. These approaches 
use the convolutional neural network (CNN) as a regressor. But, rather than attempting to reconstruct
the image from the measurements $\M y$ directly, the most successful strategies so far have been to train the CNN as a regressor between rogue initial reconstruction $\M A \by$, where $\M A: \R^M \to \R^N$, and the final, desired reconstruction \cite{jin2017deep,han2017deep}. This initial reconstruction could be obtained by FBP ($\M F\HT\by$), BP ($\HT \by$) or by any other linear operation.

Once the training is complete, the reconstruction for an unseen measurement $ \by$ is given by $\bx^* ={\cnn_{\M \btheta^*}}(\M A \by)$, where ${\cnn_{\M \btheta}}: \dimx \to \dimx$ denotes the CNN as a function and $\M \btheta^*$ denotes the internal parameters of the CNN after training.

These schemes exploit the fact that the structure of images can be learnt from representative examples. CNNs are favored because of the way they encode/represent the data in their hidden layers. In this sense, a CNN can be seen as a good prior encoder. Although the results reported so far are remarkable in terms of image quality, there is still some concern on whether or not they can be trusted, especially in the context of diagnostic imaging. The main limitation of direct algorithms such as \cite{jin2017deep} is that they do not provide any worst case performance guarantee.
Moreover, even in the case of noiseless (or low noise) measurements, there is no insurance that the reconstructed image is consistent with the measurements. This is not overly surprising because, unlike the iterative schemes, there is no feedback mechanism that imposes this minimal requirement. 

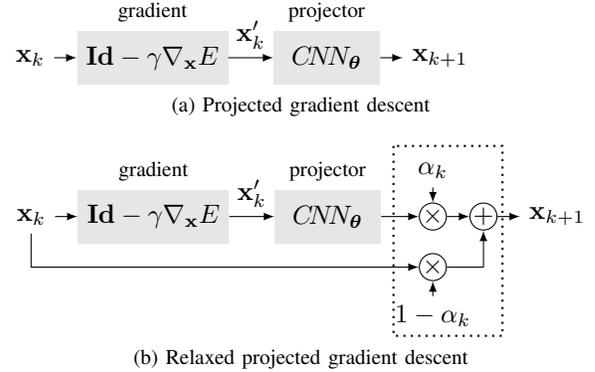
\begin{figure}
\centering
\tikzset{>=latex}

\subfloat[Projected gradient descent]{%
\begin{tikzpicture}
    \tikzstyle{block} = [fill=gray!20, rectangle, 
    minimum height=2em, minimum width=4em];
    
    \node (in) {$\M x_k$};
    \node[block, right = .3cm of in,label=above:{\footnotesize gradient}] (grad) { ${\bf{Id} -\gamma  \nabla_{\M x}}E$};
    \draw[->] (in) -- (grad);
    
    \node[block, right = .6cm of grad, label=above:{\footnotesize projector}] (CNN) {$\cnn_{\btheta}$};
    \draw[->] (grad) -- node [above] {$\M x'_k$} (CNN);
    
    \node[right = .3cm of CNN] (out) {$\M x_{k+1}$};
    \draw[->] (CNN) -- (out);
    \node[right = 1.2cm of out] {}; 
\end{tikzpicture}   
}

\subfloat[Relaxed projected gradient descent]{%
\tikzset{>=latex}
\begin{tikzpicture}
    \tikzstyle{block} = [fill=gray!20, rectangle, 
    minimum height=2em, minimum width=4em];
    
    \node (in) {$\M x_k$};
    \node[block, right = .3cm of in, label=above:{\footnotesize gradient}] (grad) { ${\bf{Id} -\gamma  \nabla_{\M x}}E$};
    \draw[->] (in) -- (grad);
    
    \node[block, right = .6cm of grad, label=above:{\footnotesize projector}] (CNN) {$\cnn_{\btheta}$};
    \draw[->] (grad) -- node [above] {$\M x'_k$} (CNN);
    
    \node[draw, circle, right = .5cm of CNN,inner sep=0pt] (timesTop) {$\times$};
    \draw[->] (CNN) -- (timesTop);
    \node[above = .2cm of timesTop] (alpha) {$\alpha_k$};   
    \draw[->] (alpha) -- (timesTop);
    
    \node[draw, circle, below = .3cm of timesTop,inner sep=0pt] (timesBot) {$\times$};
    \draw[->] (in) -- (timesBot -| in) -- (timesBot);

    \node[below = .2cm of timesBot] (1alpha) {$1-\alpha_k$};
    \draw[->] (1alpha) -- (timesBot);
    
    \node[draw, circle, right = .3cm of timesTop,inner sep=0pt] (plus) {$+$};
    \draw[->] (timesTop) -- (plus);
    \draw[->] (timesBot) -- (timesBot -| plus) -- (plus);  
    
    \node[right = .3cm of plus] (out) {$\M x_{k+1}$};
    \draw[->] (plus) -- (out);
    
  \draw[thick,dotted] ($(alpha.north west)+(-0.2,0.1)$)  rectangle ($(1alpha.south east)+(0.3,0)$);

\end{tikzpicture}   
}

    \caption{(a) Block diagram of projected gradient descent using a CNN as the projector.
    The gradient step \wrt the data-fidelity term $E=\|\bH \bx-\by\|^2$, promotes consistency with the measurements and the projector forces the solution to belong to the set of desired solutions.
    If the CNN is only an approximate projector, the scheme may diverge.
    (b)~Block diagram of the proposed relaxed projected gradient descent.
    The $\alpha_k$s are updated in such a way that the algorithm always converges (see Algorithm \ref{alg} for more details).
    \label{blockdiag}}
\end{figure}
\subsection{Contribution}
\noindent We propose a simple yet effective iterative scheme which tries to incorporate the advantages of the existing algorithms and side-steps their disadvantages (see Figure \ref{blockdiag}). Moreover, it outperforms the existing algorithms to reconstruct biomedical images from  their sparse-view CT measurements. Specifically, 

 \begin{itemize}
 \item  We initialize our reconstruction using a classical algorithm.
\item  We learn a CNN 
that acts as a projector onto a set $\Spc S$ which can be intuitively thought of as the manifold of the data (\eg, biomedical images). In this sense, our CNN encodes the prior knowledge of the data. Its purpose is to map an input image to an output image that enjoys a more structural fit to the training data than the input image. 
\item Similarly to variational methods, 
we iteratively alternate between minimizing the data-fidelity term and projecting the result onto the set $\Spc S$ by applying a suitable variant of the projected gradient descent (PGD) which ensures convergence. This scheme in spirit is similar to plug-and-play ADMM but is simpler to analyze. 
In this way, instead of performing the reconstruction by a feedbackless pipeline, we perform it by iteratively enforcing measurement consistency and injecting prior knowledge.
\end{itemize}
\subsection{Roadmap for the Paper}
The paper is organized as follows:
In Section \ref{philosophy}, we discuss the formal framework that motivates our approach. We mathematically justify the use of a projector onto a set as an effective strategy to solve inverse problems. In Section \ref{convergence}, we present an iterative algorithm inspired from PGD. It has been modified so as to converge in practical cases where the projection property is only approximate. This is crucial because, although we train the CNN as a projector using a training set, there is no guarantee that
it will act like a projector for an unseen data. 
We discuss in Section \ref{projector} a novel technique to train the CNN as a projector onto a set, especially when the training data is small (\eg, around 500 images in our case). This is followed by results and conclusion.
\subsection{Related and Prior Work}
 Deep learning has shown promising results in the cases of image denoising, superresolution and deconvolution. Recently, it has also been used to solve inverse problems in imaging using limited data \cite{jin2017deep,antholzer2017deep,han2017deep,wang2016accelerating}, and in compressed sensing \cite{Mousavi2017}. However, as discussed earlier, these regression based approaches lack the feedback mechanism that can be beneficial to solve the inverse problems. 

The other usage of deep learning is to complement the iterative algorithms. This includes learning a CNN as an unrolled version of ISTA \cite{gregor2010learning} and ADMM \cite{sun2016deep}. In \cite{adler2017solving}, inverse problems including non-linear forward models are solved by partially learning the gradient descent. In \cite{putzky2017recurrent} the iterative algorithm is replaced by a recurrent neural network (RNN). In all of these approaches the training depends entirely on the iterative scheme the neural network is used with. 

On the other hand, \cite{venkatakrishnan2013plug,chan2017plug,romano2016little,chang2017,sreehari2016plug} use plug-and-play ADMM that uses a denoiser which is learnt independently of the iterative scheme.
In \cite{chang2017}, a generative adversarial network (GAN) trained as a projector onto a set is plugged into the plug-and-play ADMM and is used to solve arbitrary linear inverse problem. However, due to the adversarial nature, the training is complicated and requires extremely large datasets (around 8 million images). Performing such training would be challenging in biomedical imaging applications because of the lack of large datasets and the high-resolution of the images ($512 \times 512$ or more). Also, in many cases, the performance of \cite{chang2017} is worse than the regression-based deep-learning methods specialized for a given inverse problem.

\section{Theoretical Framework}\label{philosophy}

The central theme of this paper is to use CNN iteratively with PGD to solve inverse problem. The task of the CNN is to act like a projector. It projects the input image to the space of desired images. To understand why this scheme will be effective, we first analyze how using a projector onto a set, combined with gradient descent, can be helpful in solving inverse problems. Proofs of all the theoretical results except Theorem \ref{thm:main} can be found in the supplementary material.
\subsection{Notation}
We consider the finite-dimensional Hilbert space $\dimx$ equipped with the scalar product $\ip{\cdot}{\cdot}$ that induces the $\ell_2$ norm $\norm{\cdot}_2$.
The spectral norm of the matrix $\bH$, denoted by $\norm{\bH}_2$, is equal to its largest singular value. For $\bx\in\dimx$ and $\varepsilon>0$, we denote by $\cB_{\varepsilon}(\bx)$ the $\ell_2$-ball centered at $\bx$ with radius $\varepsilon$, \ie,
\begin{align*}
\cB_{\varepsilon}(\bx)= \left\{\bz\in\dimx: \norm{\bz-\bx}_2\leq \varepsilon \right\}.
\end{align*} 

The operator $T:\dimx\rightarrow\dimx$ is Lipschitz-continuous with constant $L$ if
\begin{align*}
	\norm{T(\bx)-T(\bz)}_2 \leq L\norm{\bx-\bz}_2,\quad \forall \bx,\bz\in\dimx.
\end{align*}
It is contractive if it is Lipschitz-continuous with constant $L<1$ and   non-expansive if $L=1$. A fixed point $\bx^*$ of $T$ (if any) satisfies $T(\bx^{*})=\bx^*$.

Given a set $\cS\subset\dimx$, a mapping $P_{\cS}:\dimx\rightarrow\cS$ is called a projector if it satisfies the 
 {idempotent} property $P_{\cS} P_{\cS}=P_{\cS}$.
It is called an 
 {orthogonal} projector if 
\begin{align*}
P_{\cS}(\bx)=\inf_{\bz\in\cS}\norm{\bx-\bz}_2,\quad \forall\bx\in\dimx.
\end{align*}

\subsection{Constrained Least Squares}
Consider the problem of reconstructing an image $\bx\in\dimx$ from its noisy measurements $\by=\bH\bx +\bn$, where $\bH\in\dimH$ is the linear forward model and $\bn\in\dimy$ is additive white Gaussian noise. Our reconstruction  incorporates a strong form of prior knowledge about the original image: We assume that $\bx$ must lie in some set $\cS\subset\dimx$ that contains all objects of interest. The proposed way to make the reconstruction consistent with the measurements as well as with the prior knowledge is to solve the constrained least-squares problem 
\begin{align}\label{prob}
\min_{\bx\in\cS}\, \frac{1}{2}\norm{\bH\bx-\by}^2_2.
\end{align}

The condition ${\bx\in\cS}$ in \eqref{prob} plays the role of a regularizer. 
 If no two points in $\cS$ have the same measurements and incase $\by$ is noiseless, then out of all the points in $\R^N$ that are consistent with the measurement $\by$, \eqref{prob} selects a unique point $\bx^*\in\cS$. In this way, it removes the ill-posedness of the inverse problem. When the measurements are noisy,~\eqref{prob} returns a point $\bx^*\in\cS$ such that $\by^*=\bH\bx^*$ is as close as possible to $\by$. Thus, it also denoises the measurement, where the quantity $\by^*$ can be regarded as the denoised version of $\by$. 

It is remarkable that~\eqref{prob} is a generalized formulation of the regularization schemes in~\eqref{iterative}, which can be rewritten as
\begin{align*}
\min_{\bx\in\cS_R}\, \frac{1}{2}\norm{\bH\bx-\by}^2_2,
\end{align*}
where $\cS_R=\{\bx\in\dimx:R(\bx) \leq \tau\}$ for some unique $\tau$ dependent on the regularization parameter $\lambda$.



The point $\bx^{*}\in\cS$ is  called a 
 {local minimizer} of~\eqref{prob} if
\begin{align*}
\exists \varepsilon>0:\norm{\bH\bx^*-\by}_2 \leq \norm{\bH\bx-\by}_2,\forall\bx\in\cS\cap\cB_{\varepsilon}(\bx^*).
\end{align*}
\subsection{Projected Gradient Descent}
When $\cS$ is a closed convex set, it is well known~\cite{Eicke:1992} that
a solution of~\eqref{prob} can be found by PGD
\begin{align}
\bx_{k+1} &= P_{\cS} (\bx_{k} - \gamma\HTH\bx_k + \gamma\HT\by)\label{eq:PL},
\end{align}
where $\gamma$ is a step size chosen such that $\gamma < 2/\norm{\HTH}_2$. This algorithm combines the orthogonal projection onto $\cS$ with the gradient descent \wrt the quadratic objective function  (also called the Landweber update~\cite{Landweber:1951}).  PGD~\cite[Sec. 2.3]{Bertsekas:1999} is a subclass of the forward-backward splitting~\cite{CombettesW:2006,CombettesP:2011}, which is known in the $\ell_1$-minimization literature as Iterative Shrinkage/Thresholding Algorithms (ISTA)~\cite{Figueiredo2003,BectBAC:2003,daubechies2004iterative}. 

In our problem, $\cS$ is presumably non-convex, but we propose to still use the update~\eqref{eq:PL} with some projector $P_{\cS}$ that may not be orthogonal. In the rest of this section, we provide sufficient conditions on the projector $P_{\cS}$ (not on $\cS$ itself) under which~\eqref{eq:PL} leads to a local minimizer of~\eqref{prob}. Similarly to the convex case, we characterize the local minimizers of~\eqref{prob} by the fixed points of the combined operator 
\begin{align}
\G_{\gamma}(\bx) = P_{\cS} (\bx - \gamma\HTH\bx + \gamma\HT\by)\label{eq:opT}
\end{align} 
and then show that some fixed point of that operator must be reached by the iteration $\bx_{k+1}=\G_{\gamma}(\bx_k)$ as $k\rightarrow\infty$, no matter the value of $\bx_0$.
We first state a sufficient condition for each fixed point of $\G_{\gamma}$ to become a local minimizer of~\eqref{prob}.
\begin{prop}\label{thm:minimizer}
	Let $\gamma>0$ and $P_{\cS}$ be such that, for all $\bx\in\dimx$,
	\begin{align}\label{eq:local}
		\ip{\bz-P_{\cS}\bx}{\bx-P_{\cS}\bx} \leq 0, \quad  \forall  \bz\in \cS\cap \cB_{\varepsilon}(P_{\cS}\bx),
	\end{align}
	for some $\varepsilon >0$.
	Then, any fixed point of the operator $\G_{\gamma}$ in~\eqref{eq:opT} is a local minimizer of~\eqref{prob}. Furthermore, if~\eqref{eq:local} is satisfied globally, in the sense that
	\begin{align}\label{eq:global}
	\ip{\bz-P_{\cS}\bx}{\bx-P_{\cS}\bx} \leq 0, \quad \forall \bx\in\dimx,\bz \in \cS,
	\end{align}
	then any fixed point of $\G_{\gamma}$ is a solution of~\eqref{prob}. 
\end{prop}
Two remarks are in order. First, \eqref{eq:global} is a well-known property of orthogonal projections onto closed convex sets. It actually implies the convexity of $\cS$ (see ~Proposition~\ref{thm:convex}). Second, \eqref{eq:local} is much more relaxed and easily achievable, for example, as stated in Proposition~\ref{thm:union_convex}, by orthogonal projections onto unions of closed convex sets (special cases are unions of subspaces, which have found some applications in data modeling and clustering~\cite{AldroubiT:2011}).
\begin{prop}\label{thm:convex}
If $P_{\cS}$ is a projector onto $\cS\subset\dimx$ that satisfies~\eqref{eq:global}, then $\cS$ must be convex.
\end{prop}

\begin{prop}\label{thm:union_convex}
	If $\cS$ is a union of a finite number of closed convex sets in $\dimx$, then the orthogonal projector $P_{\cS}$ onto $\cS$  satisfies~\eqref{eq:local}.
\end{prop}

The above results suggest that, when $\cS$ is non-convex, the best we can hope for is to find a local minimizer of~\eqref{prob} through a fixed point of $\G_{\gamma}$. Theorem~\ref{thm:fixed_point} provides a sufficient condition for PGD to converge to a unique fixed point of $\G_{\gamma}$.

\begin{thm}\label{thm:fixed_point}
	Let   $\lambda_{\max},\lambda_{\min}$ be the largest and smallest eigenvalues of $\bH^{\T}\bH$, respectively. If $P_{\cS}$ satisfies~\eqref{eq:local} and is Lipschitz-continuous  with constant $L<(\lambda_{\max}+\lambda_{\min})/(\lambda_{\max}-\lambda_{\min})$, then, for $\gamma=2/(\lambda_{\max}+\lambda_{\min})$, the sequence $\{\bx_k\}$ generated by~\eqref{eq:PL} converges to a local minimizer of~\eqref{prob}, regardless of the initialization $\bx_0$.  
\end{thm}

It is important to note that the projector $P_{\cS}$ can never be contractive since it preserves the distance between any two points on $\cS$. Therefore, when $\bH$ has a nontrivial null space, the condition $L<(\lambda_{\max}+\lambda_{\min})/(\lambda_{\max}-\lambda_{\min})$ of Theorem~\ref{thm:fixed_point} is not feasible. The smallest possible Lipschitz constant of $P_\cS$ is $L=1$, which means $P_{\cS}$ is non-expansive. Even with this condition, it is not guaranteed that the combined operator $F_{\gamma}$ has a fixed point. This limitation can be overcome when $F_{\gamma}$ is assumed to have a nonempty set of fixed points. Indeed, we state in Theorem~\ref{thm:fixed_point2} that one of them must be  reached by iterating the 
 {averaged operator} $\alpha\id+(1-\alpha)\G_{\gamma}$, where $\alpha\in(0,1)$ and $\id$ is the identity operator. 
\begin{thm}\label{thm:fixed_point2}
Let $\lambda_{\max}$ be the largest eigenvalue of $\bH^{\T}\bH$. If $P_{\cS}$ satisfies~\eqref{eq:local} and is a non-expansive operator such that $\G_{\gamma}$ in~\eqref{eq:opT} has a fixed point for some $\gamma < 2/\lambda_{\max}$, then the sequence $\{\bx_k\}$ generated by 
	\begin{align}\label{eq:averaged}
		\bx_{k+1} = (1-\alpha)\bx_{k} + \alpha \G_{\gamma}(\bx_k),
	\end{align}
	for any $\alpha \in (0,1)$,
	converges to a local minimizer of~\eqref{prob}, regardless of the initialization $\bx_0$.  
\end{thm}

\section{Relaxation with Guaranteed Convergence}\label{convergence}
Despite their elegance, Theorems~\ref{thm:fixed_point} and~\ref{thm:fixed_point2} are not directly useful when we construct the projector $P_{\cS}$ by training a CNN, because it is unclear how to enforce the Lipschitz continuity of $P_{\cS}$ on the CNN architecture. Without putting any constraints on the CNN, however, we can still achieve the convergence of the reconstruction sequence by modifying PGD as described in Algorithm~\ref{alg}; we name it relaxed projected gradient descent (RPGD). In the proposed algorithm, the projector $P_{\cS}$ is replaced with a general nonlinear operator $\F$. We also introduce a sequence $\{c_k\}$ that governs the rate of convergence of the algorithm and a sequence $\{\alpha_k\}$ of relaxation parameters that evolves with the algorithm. The convergence of RPGD is guaranteed by Theorem~\ref{thm:main}. More importantly, if the nonlinear operator $\F$ is actually a projector and the relaxation parameters do not go all the way to 0, then RPGD converges to a meaningful point. 

\begin{algorithm}
	\caption{Relaxed projected gradient descent (RPGD)}
	\label{alg}
	\vspace*{2mm}
	\textbf{\emph{Input}}: $\bH$, $\by$, $\M A$, nonlinear operator $\F$, step size $\gamma>0$, positive sequence $\{c_n\}_{n\geq 1}$, $\bx_0=\M A \by \in\dimx$, $\alpha_0\in (0,1]$.\\
	\textbf{\emph{Output}}: reconstructions $\{\bx_k\}$, relaxation parameters $\{\alpha_k\}$.
	
	\vspace*{3mm}
	
	\hspace{2mm} $k\leftarrow 0$
	
	\hspace{3mm}{\bf while} not converged {\bf do}
	
	\hspace{6mm}$\bz_{k} = \F (\bx_{k} - \gamma\HTH\bx_k + \gamma\HT\by)$
	
	\hspace{6mm}{\bf if} $k\geq 1$ {\bf then}
	
	\hspace{9mm}{\bf if} $\norm{\bz_{{k}}-\bx_{k}}_2 > c_k\norm{\bz_{k-1}-\bx_{k-1}}_2$ {\bf then}
	
	\hspace{12mm}	$\alpha_{k} = c_k\Frac{\norm{\bz_{k-1}-\bx_{k-1}}_2}{\norm{\bz_{{k}}-\bx_{k}}_2}\,\alpha_{k-1}$
	
	\hspace{9mm}{\bf else}
		
	\hspace{12mm}	$\alpha_{k} = \alpha_{k-1}$
			
	\hspace{9mm}{\bf end if}
	 
	\hspace{6mm}{\bf end if} 
	
	\hspace{6mm}$\bx_{k+1} = (1-\alpha_k)\bx_{k} + \alpha_k \bz_{k}$
	
	\hspace{6mm}$k\leftarrow k+1$
	
	\hspace{3mm}{\bf end while}
\end{algorithm}

\begin{thm}\label{thm:main}
	 Let the input sequence $\{c_k\}$ of Algorithm~\ref{alg} be asymptotically upper-bounded by $C<1$. Then, the following statements hold true for the reconstruction sequence $\{\bx_k\}$:
	 \begin{enumerate}
	 	\item[(i)]  $\bx_k\rightarrow\bx^{*}$ as $k\rightarrow\infty$, for all choices of $\F$;
	 	\item[(ii)] if $\F$ is continuous and the relaxation parameters $\{\alpha_k\}$ are lower-bounded by $\varepsilon>0$, then $\bx^*$ is a fixed point of 
	 	\begin{align}
		 	\G_{\gamma}(\bx) = \F (\bx - \gamma\HTH\bx + \gamma\HT\by)\label{eq:opF};
	 	\end{align}
	 	\item[(iii)] if, in addition to \emph{(ii)}, $\F$ is indeed a projector onto  $\cS$ that satisfies~\eqref{eq:local}, then $\bx^*$ is a local minimizer of~\eqref{prob}.
	 \end{enumerate}
\end{thm}

We prove Theorem~\ref{thm:main} in Appendix~\ref{proof:main}. Note that the weakest statement here is (i); it always guarantees the convergence of RPGD, albeit not necessarily to a fixed point of $\G_{\gamma}$. Moreover, the assumption about the continuity of $\F$ in (ii) is automatically true when $\F$ is a CNN.

\section{Training CNN as a projector}\label{projector}
With these theoretical foundations in place, we move on to the matter of training a CNN to act as the projector in RPGD (Algorithm~\ref{alg}).
For any point $\bx \in \cS$, a projector onto $\cS$ should satisfy $P_\cS \bx = \bx$.
Moreover, we want 
\begin{equation}
\bx = P_\cS(\tilde{\bx}) ,
\end{equation}
where $\tilde{\bx}$ is any perturbed version of $\bx$.
Given a training set, $\{\bx^1,\ldots, \bx^Q\}$, of points drawn from the set $\cS$, we generate an ensemble of $N \times Q$ perturbed points, $\{ \{\tilde{\bx}^{1,1}, \ldots, \tilde{\bx}^{Q,1}\}, \ldots, \{ \tilde{\bx}^{1,N} \ldots, \tilde{\bx}^{Q,N} \}\} $ and train the CNN by minimizing the loss function
\begin{equation}\label{eq:oneMSE}
J(\btheta) = \sum_{n=1}^{N} \underbrace{\sum_{q=1}^Q\left\|\bx^q-\cnn_{\btheta}( \tilde{\bx}^{q,n})\right\|_2^2}_{J_n(\btheta)}.
\end{equation}
The optimization proceeds by stochastic gradient descent for $T$ epochs, where an epoch is defined as one pass though the training data.

It remains to select the perturbations that generate the $\bx^{q,n}$.
Our goal here is to create a diverse set of perturbations so that the CNN does not overfit one specific type.
In our experiments, while training for the $t$-th epoch, we chose
\begin{align}
\tilde{\bx}^{q,1} &= \bx^{q}&:&\text{No perturbation} \label{eq:GT} \\
\tilde{\bx}^{q,2} &=  \M A \bH \bx^{q}&:&\text{Specific linear perturbation}\label{eq:FBP}\\
\tilde{\bx}^{q,3} &= \cnn_{\btheta_{t-1}} (\tilde{\bx}^{q,2})&:&\text{Dynamic non-linear perturbation}, \label{eq:dynamic}
\end{align}
where $\M A$ is a classical linear reconstruction algorithm like FBP or BP, $\bH$ is the forward model, and $\btheta_t$ are the CNN parameters after $t$  epochs. 
We now comment on each of these perturbations in detail.

Keeping $\tilde{\bx}^{q,1}$ in the training ensemble will train the CNN with the defining property of the projector: the projector maps a point in the set $\cS$ onto itself.
If the CNN were only trained with~\eqref{eq:GT}, it would be an autoencoder~\cite{Goodfellow2016}.

To understand the perturbation $\tilde{\bx}^{q,2}$ in~\eqref{eq:FBP}, recall that $\M A \bH \bx^{q}$ is a classical linear reconstruction of $\bx^{q}$ from its measurement $\by=\bH \bx^q$.
This perturbation is useful because we will initialize RPGD with $\M A \bH \bx^{q}$.
Using only \eqref{eq:FBP} for training will return the same CNN as in \cite{jin2017deep}.

The perturbation $\tilde{\bx}^{q,3}$ in~\eqref{eq:dynamic} is the output of the CNN whose parameters $\btheta_t$ change with every epoch $t$,
thus it is a non-linear and dynamic (epoch-dependent) perturbation of $\bx^q$.
The rationale for using~\eqref{eq:dynamic} is that it greatly increases the training diversity (allowing the network to see $T$ new perturbations of each training point) without greatly increasing the total training size (only requiring an additional $Q$ gradient computations per epoch).
Moreover, ~\eqref{eq:dynamic} is in sync with the iterative scheme of RPGD, where the output of the CNN is processed with a gradient descent and is again fed into itself.

\subsection{Architecture}
The architecture we use is the same as in \cite{jin2017deep}, which is a U-net \cite{ronneberger2015u} with intrinsic skip connections among its layers and an extrinsic skip connection between the input and the output. The intrinsic skip connections help to eliminate singularities during the training \cite{orhan}. The extrinsic skip connections make this network a residual net; \ie, $\cnn=\id+ Unet$, where $\id$ denotes the identity operator and $Unet : \R^N \to \R^N$ denotes the Unet as a function. The U-net therefore actually provides the projection error (negative perturbation) that should be added to the input to get the projection.  
Residual nets have been shown to be effective in the image recognition \cite{He2016} and inverse problem cases\cite{jin2017deep}.
While the residual net architecture does not increase the capacity or the approximation power of the CNN, it does help in learning functions that are close to an identity operator, as is the case in our setting.


\subsection{Sequential Training Strategy}
We train the CNN in 3 stages. In stage 1, we train it for $T_1$ epochs \wrt the loss function $J_2$ which only uses the ensemble $\{\tilde{\bx}^{q,2}\}$ generated by~\eqref{eq:FBP}.
In stage 2, we add the ensemble $\{\tilde{\bx}^{q,3}\}$ according to~\eqref{eq:dynamic} at every epoch and then train the CNN \wrt the loss function $J_2+J_3$; we repeat this procedure for $T_2$ epochs. Finally, in stage 3, we train the CNN for $T_3$ epochs with all the ensembles $\{\tilde{\bx}^{q,1},\tilde{\bx}^{q,2},\tilde{\bx}^{q,3}\}$ to minimize the original loss function $J=J_1+J_2+J_3$ from \eqref{eq:oneMSE}. 

The above sequential procedure helps speed up the training. 
The training with $\{\tilde{\bx}^{q,1}\}$ is initially bypassed with using the residual net, which is close to the identity operator. It is only incorporated in the last few epochs of stage 3. After training with only  $\{\tilde{\bx}^{q,2}\}$ in stage 1, $\tilde{\bx}^{q,3}$ will be close to $\bx^{q}$, since it is the output of the CNN for the input $\tilde{\bx}^{q,2}$. This will ease the training with $\{\tilde{\bx}^{q,3}\}$, which is added after stage 1.

\section{Experiments}
\label{ctreconstruction}

We validate our proposed method on the difficult case of sparse-view CT reconstruction with low dosage exposure. 
The measurement operator $\M H$ is now the Radon transform. It maps an image to the values of its integrals along a known set of lines~\cite{kak}.
In 2D, these measurements can be indexed by the angles and offsets of the lines and arranged in a 2D 
{sinogram}.
We are particularly interested in the case where the total number of measurements is smaller than the number of pixels in the reconstruction.
For example, we aim to reconstruct a (512 $\times$ 512) image from 45 angles, each with 729 offsets sinogram; \ie, to reconstruct $\bx \in \R^{512 \times 512}$ from $\by \in \R^{45 \times 729}$. This corresponds to about 8 times fewer measurements than the image to be reconstructed.
\subsection{Dataset}
Our dataset consists of clinically realistic invivo ($512 \times 512$) CT scans of human abdomen from Mayo clinic for the AAPM  Low  Dose  CT  Grand  Challenge \cite{Lowdose}. This data includes CT scans of 10 patients obtained using full dose. We use 475 images from 9 patients for training and 25 images from the other patient for testing. 
This is the same data used in~\cite{jin2017deep}.
These images serve as the ground truth.

From these images, we generate the measurements (sinograms) using the \texttt{radon} command in Matlab, which corresponds to the forward model $\bH$. The sinograms always have 729 offsets per view, but we vary the number of views in different experiments. Our task is to reconstruct these images from their sparse-view sinograms.
We take 2 scenarios: 144 views and 45 views, which corresponds to $\small{\times}$5 and $\small{\times}$16 dosage reductions (assuming a full-view sinogram has 720 views).

The backprojection $\HT$ is implemented via the $\texttt{iradon}$ command with a normalization to satisfy the adjoint property. 
To make the experiments more realistic and to reduce the inverse crime, the sinograms are generated by perturbing the angles of the views slightly by adding a zero-mean additive white Gaussian noise (AWGN) with standard deviation of 0.05 degrees. This creates a slight mismatch between the actual measurement process and the forward model $\bH$. We also add various amounts of zero-mean Gaussian noise to the sinograms. The SNR of the sinogram $\by+\bn$ is defined as
\begin{equation}\label{eq:ySNR}
 \text{SNR}(\by+\M n,\by) = 20 \log_{10}\left(\Frac{\norm{\by}_2}{\norm{\bn}_2}\right).
\end{equation}

Given the ground truth $\bx$, our figure of merit for the reconstructed $\bx^*$ is the regressed SNR, given by
\begin{equation}
\text{SNR} (\bx^*,\bx) =\arg\max_{a,  b} \text{SNR}(a \M x^* +b, \bx),
\end{equation}
where the scalars $a$ and $b$ serve to scale the data and remove any DC offset, which can greatly affect the SNR but are of little practical importance.
   
\subsection{Comparison Methods}
We compare four reconstruction methods and report the SNRs for all of them.
\begin{enumerate}
    \item \textbf{FBP} is the classical direct inversion of the Radon transform $\bH$, here implemented in Matlab by the \texttt{iradon} command with the \texttt{ram-lak} filter and \texttt{linear} interpolation as options.
    \item \textbf{TV} solves 
    \begin{equation}
    \min_{\M x} \left(\half\|\M H\M x-\M y\|_2^2+\lambda \|\M x\|_{\text TV}\right) \, \text{s.t. } \bx >0.
    \end{equation}
     The optimization is carried out via ADMM~\cite{boyd2011distributed}. For a given testing image the parameter $\lambda$ is tuned so as to maximize the SNR of the reconstruction.
    \item \textbf{FBPconv} is the deep-learning-based regression technique \cite{jin2017deep} that corresponds to a CNN trained with only the ensemble in~\eqref{eq:FBP}. In the testing phase, the FBP of the measurements 
    is fed into the trained CNN to output the reconstruction image. 
    \item \textbf{RPGD} is our proposed method which is described in Algorithm~\ref{alg} where the nonlinear operator $F$ is the CNN trained as a projector (as discussed in section \ref{projector}).
\end{enumerate}

\subsection{Training and Selection of Parameters}
We now describe how training and/or parameter selection occurred for the  reconstruction methods.
FBP has no free hyperparameters.
For TV, we chose $\lambda$ by a grid search through 20 values for each test image. While carrying the optimization with ADMM we put the penalty term, $\rho = \lambda$. 
The rationale for this heuristic is that the soft-threshold parameter is of the same order of magnitude as the image gradients.
We set the number of iterations to 100, which was enough to show good empirical convergence.
 

As discussed in section \ref{projector}, the CNNs for RPGD is trained in 3 stages, with the following configurations:
\begin{itemize}
    \item $\small{\times}$5, no noise: $T_1=80$, $T_2=49$, $T_3=5$.
    \item $\small{\times}$16, no noise: $T_1=71$, $T_2=41$, $T_3=11$.
\end{itemize}
We used the CNN obtained right after the first stage for FBPconv, since during this stage only the training ensemble in \eqref{eq:FBP} was taken into account. We empirically found that the training error $J_2$ converged in $T_1$ epochs of stage 1, yielding an optimal performance for FBPconv.

In addition, we also trained the CNNs \wrt 40 dB noise level in the measurements, by replacing the ensemble in~\eqref{eq:FBP} with $\{\M A \by^q\}$ where $\by^q=\bH \bx^q+\bn$. With 20\% probability, we also perturbed the views of the measurements with AWGN of 0.05 standard deviation so as to enforce robustness to model mismatch. 
These CNNs were initialized with the ones obtained after the first stage of the noiseless training and were then trained with the following configurations:
\begin{itemize}
 \item $\small{\times}$5, 40-dB noise: $T_1=35$, $T_2=49$, $T_3=5$.
    \item $\small{\times}$16, 40-dB noise: $T_1=32$, $T_2=41$, $T_3=11$.
\end{itemize}
Similarly to the noiseless case, the CNNs obtained after the first and the third stage of the above training were used in FBPconv and RPGD, respectively. For clarity, these variants will be referred to as \textbf{FBPconv40} and $\textbf{RPGD40}$.

During the training, the learning rate was logarithmically decreased  from $10^{-2}$ to $10^{-3}$ in stage 1 and kept at $10^{-3}$ for stages 2 and 3. The batch size was fixed to 2, the gradient above $10^{-2}$ was clipped and the momentum was set to $0.99$. The total training time for the noiseless case was around 21.5 hours on GPU Titan X (Pascal architecture). 

The hyper-parameters for RPGD were chosen as follows: The relaxation parameter $\alpha_0$ was initialized with 1, the sequence $\{c_k\}$ was set to a constant $C$ where $C=0.99$ for RPGD and $C = 0.8$ for RPGD40. 
For each noise and dosage reduction level, the only free parameter, $\gamma$, was tuned to give the best average SNR over 25 test images. In all experiments, the gradient step was removed from the first iteration. On the GPU, one iteration of RPGD takes less than 1 second. The algorithm is stopped when the residual $\|\bx_{k+1}-\bx_k\|_2$ reaches a value less than 1, which is sufficiently small compared to the dynamic range [0,\,350] of the image. It takes around 1-2 minutes to reconstruct an image with RPGD.

 \section{Results}\label{results}
\begin{table}[t]
\centering\caption{ Averaged reconstruction SNRs over 25 images for 16 and 5 times dosage reductions with low measurement noise. \label{tab:lownoise}}
\label{my-label}
\begin{tabular}{llcccc}
\hline\hline
\multicolumn{2}{l|}{\begin{tabular}[c]{@{}l@{}}Measurement\end{tabular}} & \multicolumn{4}{l}{~~~~~~~~~~~~~Methods}    \\ \cline{3-6} 
\multicolumn{2}{l|}{SNR (dB)  $\downarrow$}                                                 & FBP & TV & FBPconv & {RPGD} \\ \hline
\multicolumn{1}{l|}{\multirow{2}{*}{$\small{\times}$16}}   & \multicolumn{1}{l|}{$\infty$}   &   12.74  &  24.21  &  26.19 &  \textbf{27.02}  \\
\multicolumn{1}{l|}{}   & \multicolumn{1}{l|}{70}           &12.73      & 24.20    &  26.18    &  \textbf{26.94}     \\ \hline
\multicolumn{1}{l|}{\multirow{2}{*}{$\small{\times}$5}} & \multicolumn{1}{l|}{$\infty$}  &  24.19   &  {30.80}  &   32.09  &   \textbf{32.62}  \\
\multicolumn{1}{l|}{}   & \multicolumn{1}{l|}{70}  &  24.15   & 30.74   & {32.08} & \textbf{32.56} \\
 \hline\hline                                              
\end{tabular}
\end{table}
\begin{table}[h]
\centering
\caption{ Averaged reconstruction SNRs over 25 images for 16 and 5 times dosage reductions with high measurement noise.\label{tab:highnoise}}
\label{my-label}
\begin{tabular}{llcccc}
\hline\hline
\multicolumn{2}{l|}{\begin{tabular}[c]{@{}l@{}}Measurement\end{tabular}} & \multicolumn{4}{l}{~~~~~~~~~~~~~Methods}    \\ \cline{3-6} 
\multicolumn{2}{l|}{SNR (dB) $ \downarrow$}                                                 & FBP & TV & FBPconv40 & {RPGD40} \\ \hline
\multicolumn{1}{l|}{\multirow{3}{*}{$\small{\times}$16}}   & \multicolumn{1}{l|}{45}   &   11.08  &  22.59  &  20.87 &  \textbf{24.16}  \\
\multicolumn{1}{l|}{}   & \multicolumn{1}{l|}{40}           &\phantom{0}9.09      & 21.45    &  23.26    &  \textbf{23.73}     \\
\multicolumn{1}{l|}{}   & \multicolumn{1}{l|}{35}          & \phantom{0}6.51    & 20.01   & 16.20     &    \textbf{22.59} \\ \hline
\multicolumn{1}{l|}{\multirow{3}{*}{$\small{\times}$5}} & \multicolumn{1}{l|}{45}  &  18.85   &  \textbf{27.18}  &   22.56  &   \textbf{27.17}  \\
\multicolumn{1}{l|}{}   & \multicolumn{1}{l|}{40}  &  14.96   & 25.46   &  \textbf{28.24} &  {27.61} \\
\multicolumn{1}{l|}{}   & \multicolumn{1}{l|}{35}  & 10.76    & 23.44   & 18.90  &  \textbf{24.58} \\ \hline\hline                        \end{tabular}
\end{table}
 \begin{figure*}[htbp]
 \includegraphics[width=\linewidth,trim=0mm 0mm 0mm 0mm, 
  clip=true]{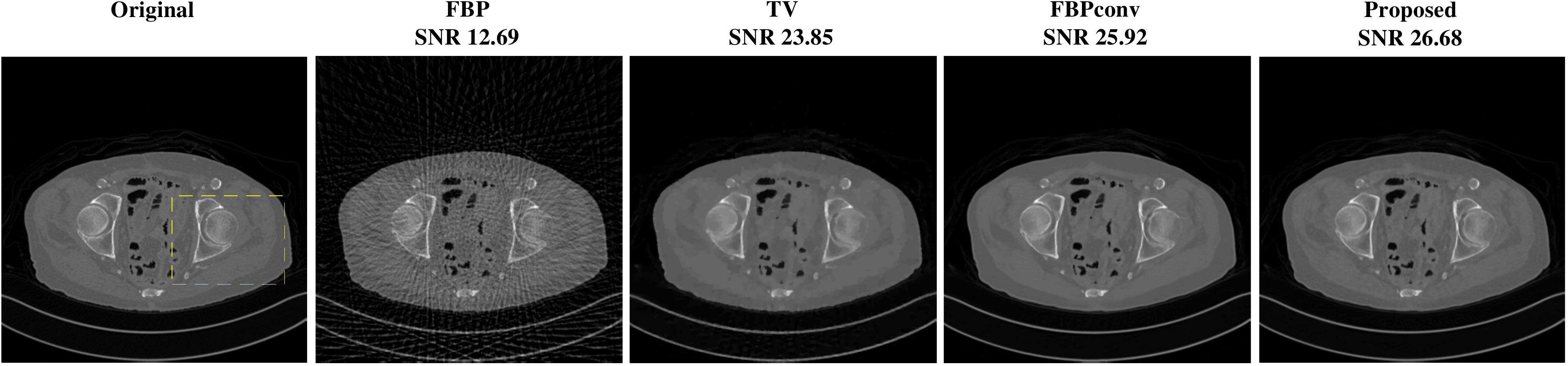}
 \includegraphics[width=\linewidth,height=3cm ]{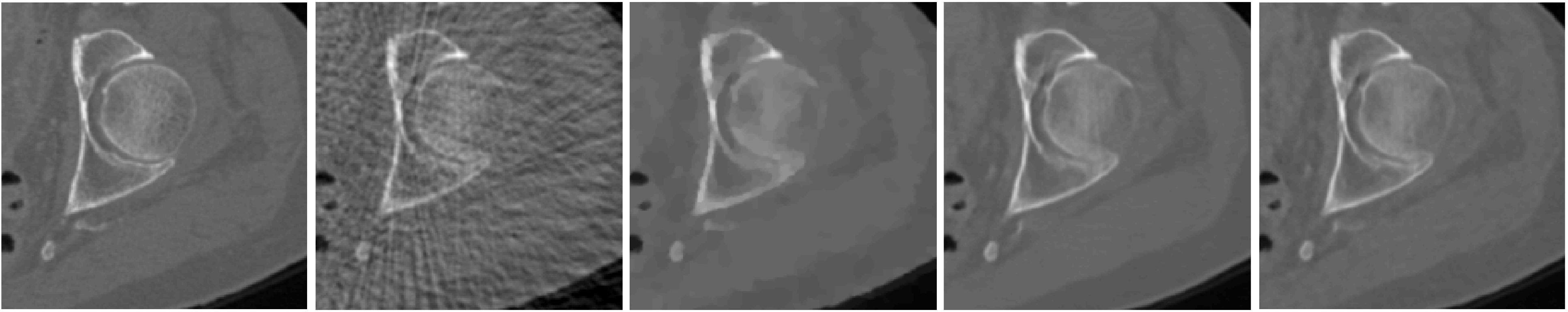}
 \centerline{(a)}\vspace{2mm}
 \includegraphics[width=\linewidth,height=3.3cm ]{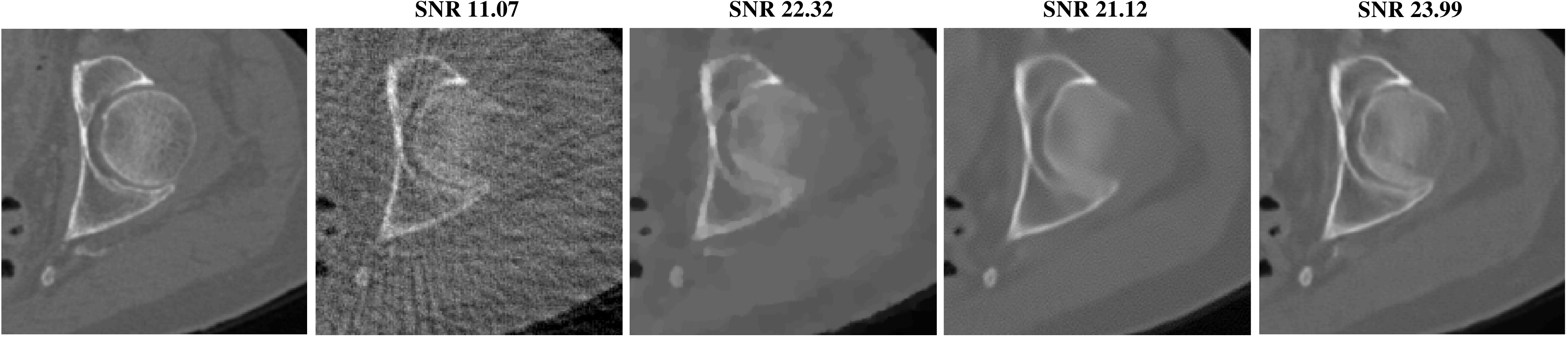}\vspace{2mm}
 \includegraphics[width=\linewidth,height=3.3cm ]{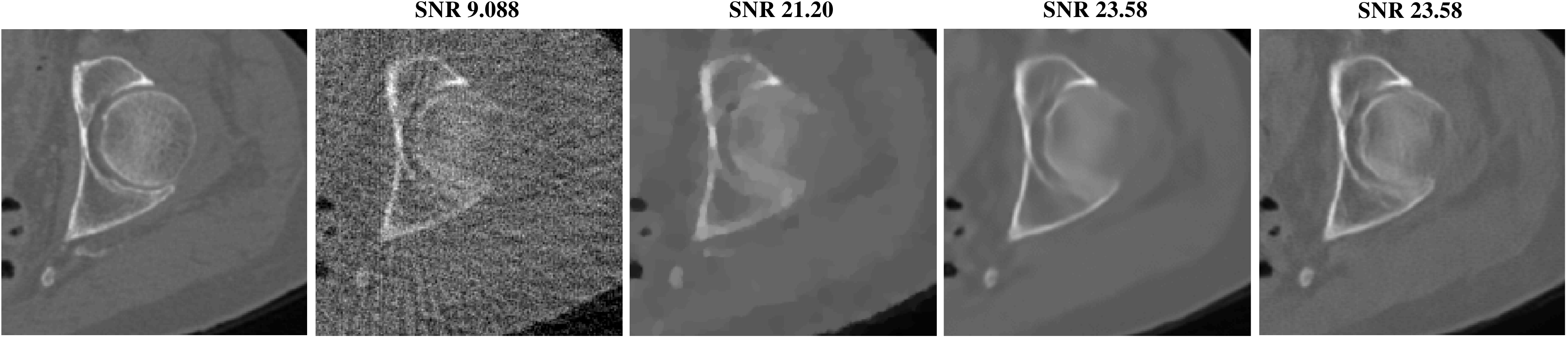}
  \centerline{(b)}
\caption{ (a) Reconstructions of a test image from noiseless measurements with 45 views ($\small{\times}$16 dosage reduction) using FBP, TV, FBPconv, and RPGD (proposed): first row shows the full images, second row shows their zoomed versions. (b) Zoomed reconstructions of the same image when the measurement is noisy with 45 dB (first row) and 40 dB (second row) SNR. In these cases, FBPconv and RPGD are replaced by FBPconv40 and RPGD40, respectively.} 
\label{fig:x16image}
\end{figure*}

 \begin{figure*}[htbp]
 \includegraphics[width=\linewidth,trim=0mm 0mm 0mm 0mm, 
  clip=true,height=4cm]{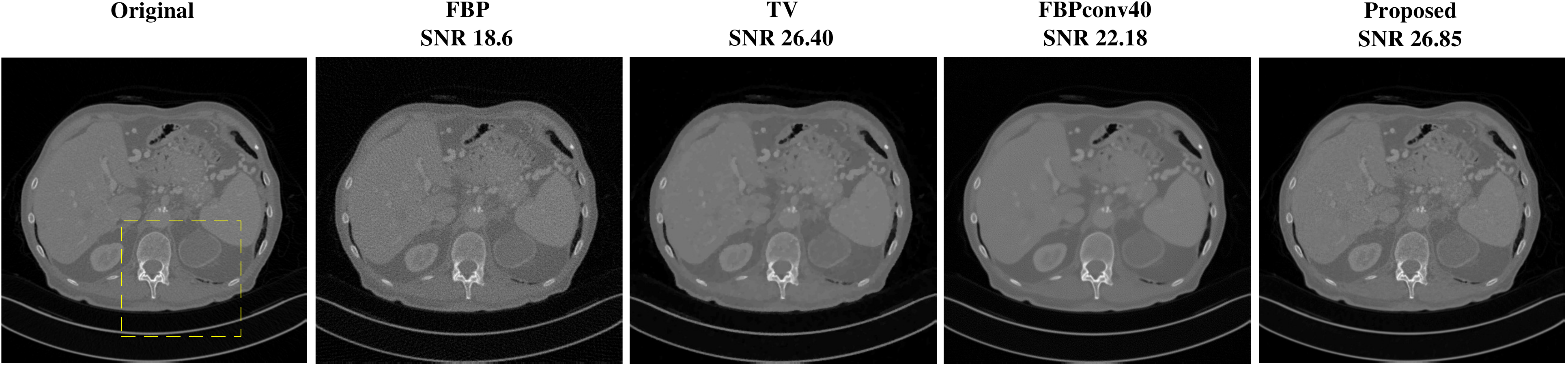}
 \includegraphics[width=\linewidth,height=3cm ]{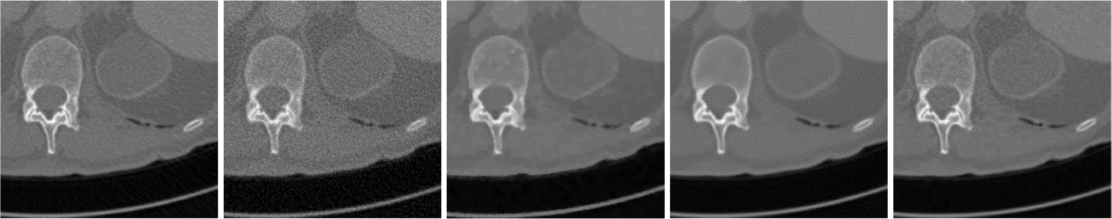}\\

  \includegraphics[width=\linewidth,trim=0mm 0mm 0mm 1mm,clip=true,height=3.3cm ]{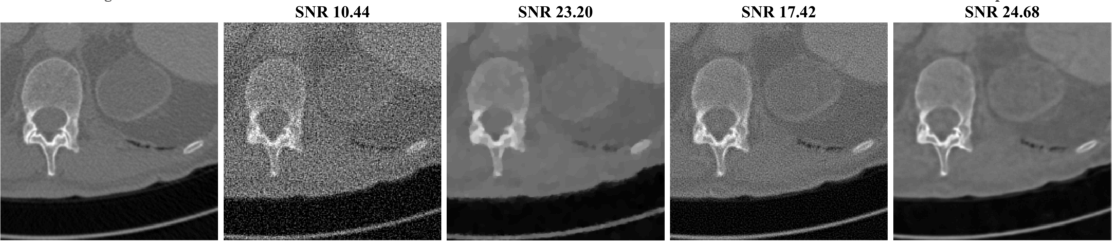}
\caption{ Reconstructions of a test image from noisy measurements with 144 views ($\small{\times}$5 dosage reduction) using FBP, TV, FBPconv40, and RPGD40 (proposed). First row shows the results for measurement noise with SNR = 45 dB; second row shows their zoomed version. Third row shows the zoomed results for measurement noise with SNR = 35 dB. 
} 
\label{fig:x5image}
\end{figure*}

\begin{figure*}
\begin{minipage}[b]{0.33\linewidth}
  \centering
  \centerline{ \includegraphics[width=\linewidth, trim={0 0 0 0cm },clip,height=5cm]{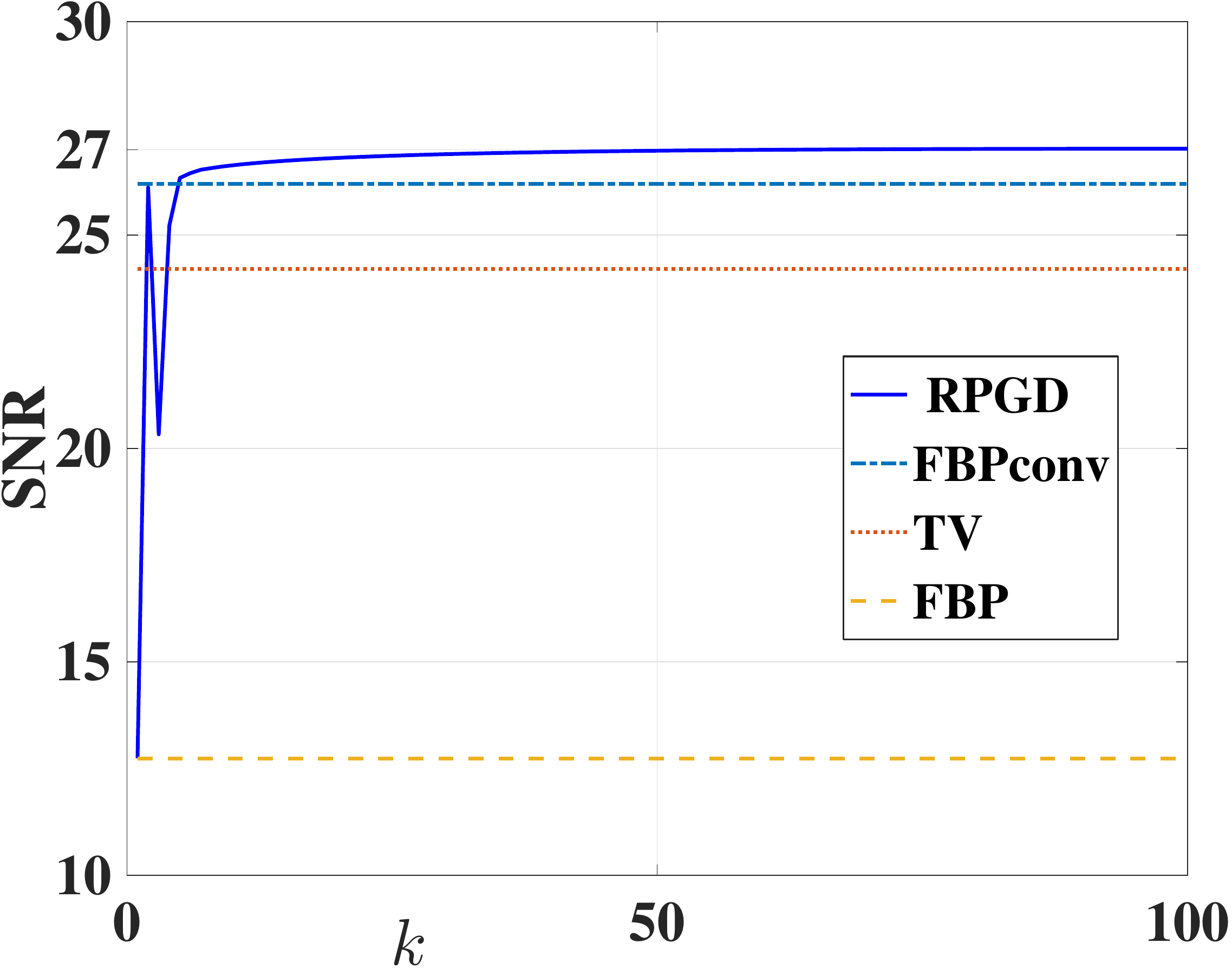}}
  \centerline{(a)}
\end{minipage}
\begin{minipage}[b]{0.33\linewidth}
  \centering
  \centerline{\includegraphics[width=\linewidth, trim={0 0 0 0cm },clip,height=5cm]{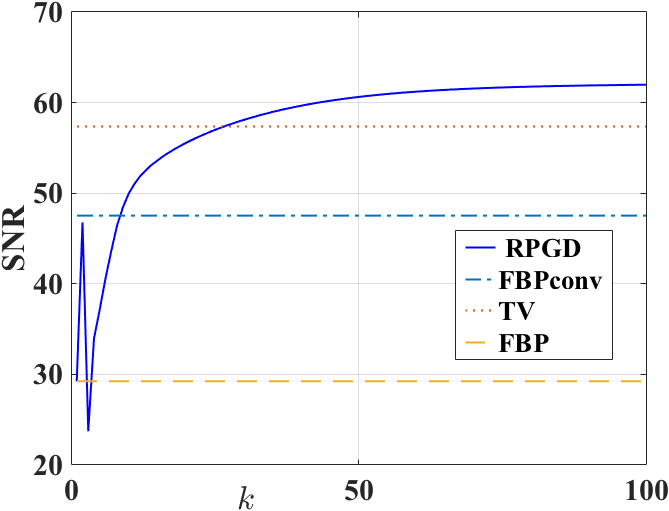}}
  \centerline{(b)}
\end{minipage}
\begin{minipage}[b]{0.33\linewidth}
  \centering
  \centerline{\includegraphics[width=\linewidth,height=5cm ]{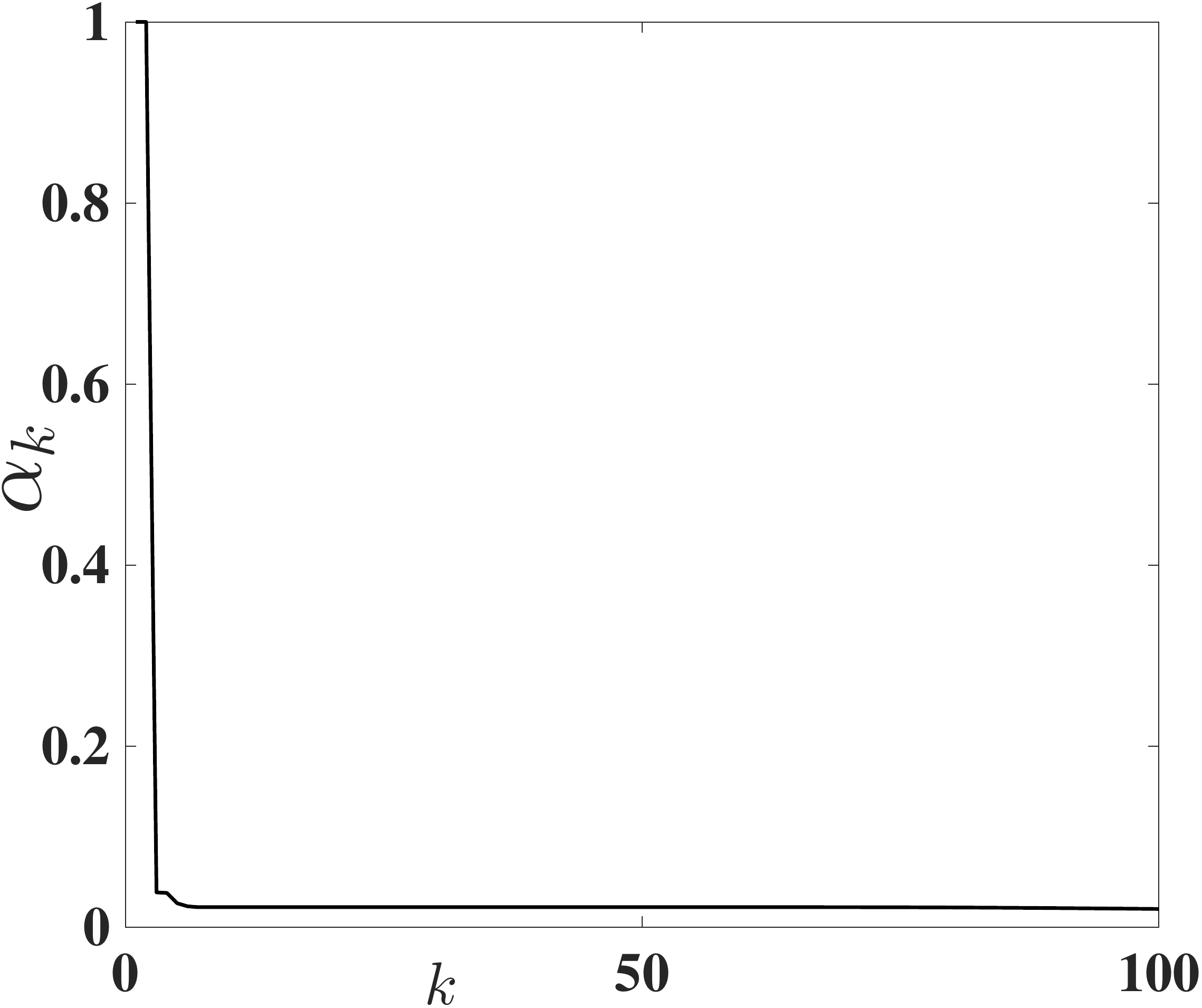}}
  \centerline{(c)}
\end{minipage}
\caption{Convergence with iteration $k$ of RPGD for the $\small{\times}$16, no-noise case when $C=0.99$. Results are averaged over 25 test images. (a)  SNRs of $\bx_k$ \wrt the ground-truth image. (b) SNRs of $\bH \bx_k$ \wrt the ground-truth sinogram. (c) Evolution of the relaxation parameters $\alpha_k$.}\label{fig:convergence}
\end{figure*}





Tables \ref{tab:lownoise} and \ref{tab:highnoise} report the results of various methods for low and high measurement noise, respectively. FBPconv and RPGD are used for low noise, while FBPconv40 and RPGD40 are used for high noise. The reconstruction SNRs are averaged over the 25 test images. 

In the low-noise cases (Table~\ref{tab:lownoise}), the proposed method, RPGD, outperforms all the others for both $\small{\times}5$ and $\small{\times}16$ reductions. FBP performs the worst but is able to retain enough information to be utilized by FBPConv and RPGD. Thanks to the convexity of the iterative scheme, TV is able to perform well but tends to smooth textures and edges. On the other hand, 
FBPConv outperforms TV. However, it is outperformed by RPGD. This is mainly due to the feedback mechanism in RPGD which lets RPGD use the information in the given measurements to increase the quality of the reconstruction. In fact, for the $\small{\times}16$, no noise case, the SNRs of the sinogram of the reconstructed images for TV, FBPconv, and RPGD are around 47 dB, 57 dB, and 62 dB, respectively. This means that not only reconstruction using RPGD has better image quality but is also more reliable since it is consistent with the given noiseless measurement. 

In the noisier cases (Table \ref{tab:highnoise}), RPGD40 outperforms the other methods in low-view cases ($\small{\times}16$) and is more consistent in performance than the others in high-view ($\small{\times}5$) cases. FBPconv40 substantially outperforms TV in the two scenarios with 40-dB noise measurement, over which it was actually trained. 
However, as the level of noise deviates from 40 dB, the performance of FBPconv40 degrades significantly.  Surprisingly, its performances in the 45-dB cases are much worse than those in the corresponding 40-dB cases.
This implies that FBPConv is highly sensitive to the difference between the training and the testing conditions. By contrast,
RPGD40 is more robust to this difference due to its iterative correction.
In fact, for $\small{\times} 16$ case with 45-dB and 35-dB noise level, it outperforms FBPconv40 by around 3.5 dB and 6 dB, respectively.

Fig.~\ref{fig:x16image} (a) illustrates the reconstructions of a test image for $\small{\times} 16$ case when measurement is noiseless.
FBP is dominated by line artifacts, while TV satisfactorily removes these but blurs the fine structures. FBPConv and RPGD, on the other hand, are able to reconstruct these details. The zoomed version shows that RPGD is able to reconstruct the fine details better than the others. This observation remains the same when the measurement quality degrades. Fig.~\ref{fig:x16image} (b) shows the reconstructions for 45-dB and 40-dB noise levels. In these scenarios, RPGD40 is significantly better than both FBPconv40 and TV.  

Fig.~\ref{fig:x5image} compares the reconstructions for the $ \small{\times} 5$ case when the noise levels are 45 dB and 35 dB. It is visible that FBPconv40 results in a noisy image and TV is again blurred. RPGD40 retains the fine structures and is the best performer.

\subsection{Convergence of RPGD}
 Figs.~\ref{fig:convergence} (a) and (b) shows the evolution of SNR of images $\bx_k$ and their measurements $\M H \bx_k$ \wrt the ground truth image and ground truth measurement, respectively. Fig.~\ref{fig:convergence} (c) shows the $\alpha_k$ \wrt the iteration $k$.
 The results are averaged over 25 test images for $\small{\times}16$, no noise case and $C=0.99$.
 RPGD outperforms all the other methods in terms of both image quality and measurement consistency.
 
 Due to the high value of the step size ($\gamma=2 \times 10^{-3}$) and the large difference $\bH \bx_k -\by$, the initial few iterations have large gradients resulting in the instability of the algorithm. The reason is that the CNN is fed with $\bx_k -\gamma \HT(\bH \bx_k -\by)$ which is drastically different from the perturbations on which it was trained.
 In this situation, $\alpha_k$ decreases steeply and stabilizes the algorithm. 
At convergence, $\alpha_k\neq 0$, therefore, according to Theorem~\ref{thm:main}, $\bx_{100}$ is the fixed point of \eqref{eq:opF} where $F=\cnn$.
 

 \section{Conclusion}
We have proposed a simple yet effective iterative scheme (RPGD) where one step of enforcing measurement consistency is followed by a CNN which tries to project the solution onto the set of the data that we are interested in. 
The whole scheme is ensured to be convergent. We also introduced a novel method to train a CNN that acts like a projector using a reasonably small dataset (475 images).
For sparse-view CT reconstruction our method outperforms the previous techniques for both noiseless and noisy measurements. 

The proposed framework can be used to solve other inverse problems like super-resolution, deconvolution, accelerated MRI, \etc This can bring more robustness and reliability to the current deep learning based techniques.

\begin{appendix}
\subsection{Proof of Theorem~\ref{thm:main}} \label{proof:main}
	(i)
	Set $\br_k=\bx_{k+1}-\bx_k$. On one hand, it is clear that
	\begin{align}
		\br_k &= (1-\alpha_k)\bx_k + \alpha_k \bz_k - \bx_{k}= \alpha_k\left(\bz_k-\bx_k\right).
	\end{align}
	On the other hand, from the construction of $\{\alpha_k\}$,
	\begin{align}
	&\alpha_{k}\norm{\bz_{k} - \bx_{k}}_2  \leq c_k\alpha_{k-1}\norm{\bz_{k-1} - \bx_{k-1}}_2\nonumber\\
	\Leftrightarrow &\quad\quad\quad\ \,\norm{\br_{k}}_2 \leq c_k \norm{\br_{k-1}}_2.\label{eq:inequality}
	\end{align}
	Iterating~\eqref{eq:inequality} gives
	 \begin{align}\label{eq:key_bound}
	 \norm{\br_{k}}_2 \leq \norm{\br_0}_2 \prod_{i=1}^{k} c_{i}, \quad \forall k\geq 1.
	 \end{align}
	We now show that $\{\bx_k\}$ is a Cauchy sequence. Since $\{c_k\}$ is asymptotically upper-bounded by $C<1$, there exists $K$ such that $c_k \leq C,\forall k> K$. Let $m,n$ be two integers such that $m>n>K$. By using~\eqref{eq:key_bound} and the triangle inequality, 
	\begin{align}
		\norm{\bx_m-\bx_n}_2 &\leq \sum_{k=n}^{m-1} \norm{\br_{k}}_2\leq \norm{\br_0}_2 \prod_{i=1}^{K} c_{i} \sum_{k=n-K}^{m-1-K} C^k\nonumber\\
		&\leq \left(\norm{\br_0}_2 \prod_{i=1}^{K} c_{i}\right) \frac{C^{n-K}-C^{m-K}}{1-C}.
	\end{align}
	The last inequality proves that $\norm{\bx_m-\bx_n}_2\rightarrow 0$ as $m\rightarrow\infty,n\rightarrow\infty$, or $\{\bx_k\}$ is a Cauchy sequence in the complete metric space $\dimx$. As a consequence, $\{\bx_{k}\}$ must converge to some point $\bx^*\in\dimx$. 
	
	(ii) Assume from now on that $\{\alpha_k\}$ is lower-bounded by $\varepsilon>0$. By definition, $\{\alpha_k\}$ is also non-increasing and, thus, convergent to $\alpha^*>0$. Next, we rewrite the update of $\bx_k$ in Algorithm~\ref{alg} as
	\begin{align}\label{eq:to_take_limit}
		\bx_{k+1} = (1-\alpha_k)\bx_{k} + \alpha_k \G_{\gamma}(\bx_{k}),
	\end{align}
	where $\G_{\gamma}$ is defined by~\eqref{eq:opF}. Taking the limit of both sides of~\eqref{eq:to_take_limit} leads to 
	\begin{align}\label{eq:limit}
	\bx^* &= (1-\alpha^*)\bx^* + \alpha^*\lim_{k\rightarrow\infty}\G_{\gamma}(\bx_k).
	\end{align}
	Moreover, since the nonlinear operator $\F$ is continuous, $\G_{\gamma}$ is also continuous. Hence,
	\begin{align}\label{eq:continuous}
		\lim_{k\rightarrow\infty}\G_{\gamma}(\bx_k) =\G_{\gamma}\left(\lim_{k\rightarrow\infty}\bx_k\right) = \G_{\gamma}(\bx^*). 
	\end{align}
	By plugging~\eqref{eq:continuous} into~\eqref{eq:limit}, we get that $\bx^*=\G_{\gamma}(\bx^*)$, which means $\bx^*$ is a fixed point of the operator $\G_{\gamma}$. 
	
	(iii)
	Now that $\F = P_{\cS}$ satisfies~\eqref{eq:local}, we invoke Proposition~\ref{thm:minimizer} to infer that $\bx^*$ is a local minimizer of~\eqref{prob}, thus completing the proof. 
\end{appendix}
\section{Acknowledgment}
The authors would like to thank Emmanuel Soubies for his helpful suggestions on training the CNN.
We thankfully acknowledge  the  support  of  NVIDIA  Corporation
which donated the Titan X GPU used in this research. The authors would like to thank Dr. Cynthia McCollough, the Mayo Clinic, the American Association of Physicists in Medicine, and grants EB017095 and EB017185 from the National Institute of Biomedical Imaging and Bioengineering for giving opportunities to use real-invivo CT DICOM images.

\newpage


 \section{Supplementary material}
 \subsection{Proof of Proposition~\ref{thm:minimizer}} \label{proof:minimizer}
 Suppose that~\eqref{eq:local} is fulfilled and let $\bx^*\in\cS$ be a fixed point of $\G_{\gamma}$. We show that $\bx^*$ is also a local minimizer of~\eqref{prob}. Indeed,
 setting $\bx=\bx^*-\gamma\HTH\bx^{*} + \gamma\HT\by$ leads to $P_{\cS}\bx=\bx^*$. Then, there exists $\varepsilon>0$ such that, for all $\bz\in\cS\cap\cB_{\varepsilon}(\bx^{*})$, 
 \begin{align*}
 0&\geq\ip{\bz-P_{\cS}\bx}{\bx-P_{\cS}\bx}\\
 &=\gamma\ip{\bz-\bx^*}{\HT\by-\bH^{\T}\bH\bx^{*}}\\
 &=\frac{\gamma}{2}\left(\norm{\bH\bx^*-\by}^2_2 -\norm{\bH\bz-\by}^2_2 + \norm{\bH\bx^*-\bH\bz}^2_2\right).\label{eq:nonnegativity}
 \end{align*}
 Since $\gamma>0$, the last inequality implies that 
 \begin{align*}
 \norm{\bH\bx^*-\by}^2_2 \leq \norm{\bH\bz-\by}^2_2,\quad\forall \bz\in\cS\cap\cB_{\varepsilon}(\bx^{*}),
 \end{align*}
 which means that $\bx^{*}$ is a local minimizer of~\eqref{prob}.

 Assume now that $P_{\cS}$ satisfies~\eqref{eq:global}. By just removing the $\varepsilon$-ball in the above argument, one easily verifies that
 \begin{align*}
 \norm{\bH\bx^*-\by}^2_2 \leq \norm{\bH\bz-\by}^2_2,\quad\forall \bz\in\cS,
 \end{align*}
 which means that $\bx^*$ is a solution of~\eqref{prob}.	
 \subsection{Proof of Proposition~\ref{thm:convex}} \label{proof:convex}
 We prove by contradiction. Assuming that $\cS$ is non-convex, there must exist $\bx_1,\bx_2\in\cS$ and $\alpha\in (0,1)$ such that $\bx=\alpha\bx_1 + (1-\alpha)\bx_2\notin \cS$. Since $P_{\cS}\bx\in\cS$, it must be that 
 \begin{align*}
 0 &<\norm{\bx-P_{\cS}\bx}^2_2=\ip{\bx-P_{\cS}\bx}{\bx-P_{\cS}\bx}\\
 &=\alpha\ip{\bx_1-P_{\cS}\bx}{\bx-P_{\cS}\bx} \\
 &\quad+ (1-\alpha)\ip{\bx_2-P_{\cS}\bx}{\bx-P_{\cS}\bx}.
 \end{align*}
 Thus, there exists $i\in\{0,1\}$ such that 
 \begin{align*}
 \ip{\bx_i-P_{\cS}\bx}{\bx-P_{\cS}\bx} > 0,
 \end{align*}
 which violates~\eqref{eq:global}. So, $\cS$ is convex.

 \subsection{Proof of Proposition~\ref{thm:union_convex}} \label{proof:union_convex}
 Suppose that $\cS=\bigcup_{i=1}^{n}\cC_i$, where $\cC_i$ is a closed convex set for all $i=1,\ldots,n$. The statement is trivial when $n=1$; assume now that $n\geq 2$. Let $\bx\in\dimx$ and $\hat{\bx}$ be the orthogonal projection of $\bx$ onto $\cS$. Consider two cases.

 \emph{Case 1}: $\hat{\bx}\in\bigcap_{i=1}^n\cC_i$.\\
 It is then clear that 
 \begin{align*}
 \norm{\hat{\bx}-\bx}_2 \leq \norm{\bz-\bx}_2,\forall \bz\in\cC_i,\forall i.
 \end{align*}
 This means that $\hat{\bx}$ is the orthogonal projection of $\bx$ onto each $\cC_i$. Consequently,
 \begin{align*}
 \ip{\bz-\hat{\bx}}{\bx-\hat{\bx}} \leq 0,\forall \bz\in\cC_i,\forall i\leq n,
 \end{align*}
 which implies that~\eqref{eq:local} holds true for all $\varepsilon > 0$.

 \emph{Case 2}: $\hat{\bx}\notin\bigcap_{i=1}^n\cC_i$.\\
 Without loss of generality, there exists $m<n$ such that 
 \begin{align}
 \hat{\bx}\in \bigcap_{i=1}^m \cC_i,\quad \hat{\bx}\notin \bigcup_{i=m+1}^n \cC_i.
 \end{align} 
 Let $d$ be the distance from $\hat{\bx}$ to the set $\cT=\bigcup_{i=m+1}^n \cC_i$. Since each $\cC_i$ is closed, $\cT$ must be closed too and, so, $d>0$. We now choose $0<\varepsilon<d$. Then, $\cB_\varepsilon(\hat{\bx})\cap\cT=\emptyset$. Therefore,
 \begin{align}\label{eq:intersect}
 \cS\cap\cB_\varepsilon(\hat{\bx}) =\bigcup_{i=1}^{m}\left(\cC_i\cap \cB_\varepsilon(\hat{\bx})\right) = \bigcup_{i=1}^{m}\tilde{\cC}_i,
 \end{align}
 where $\tilde{\cC}_i=\cC_i\cap \cB_\varepsilon(\hat{\bx})$ is clearly a convex set, for all $i\leq m$. It is straightforward that $\hat{\bx}$ is the orthogonal projection of $\bx$ onto the set $\bigcup_{i=1}^{m}\tilde{\cC}_i$  and that $\bx\in\bigcap_{i=1}^{m}\tilde{\cC}_i$. We are back to Case 1 and, therefore,
 \begin{align}\label{eq:case}
 \ip{\bz-\hat{\bx}}{\bx-\hat{\bx}} \leq 0,\forall \bz\in\tilde{\cC}_i,\forall i\leq m.
 \end{align}
 From~\eqref{eq:intersect} and~\eqref{eq:case}, \eqref{eq:local} is fulfilled for the chosen $\varepsilon$.
 \subsection{Proof of Theorem~\ref{thm:fixed_point}} \label{proof:fixed-point}
 	Let $\{\lambda_i\}$ denote the set of eigenvalues of $\HTH$. We first have that, for all $\bx\in\dimx$, 
 	\begin{align}\label{eq:gradient_bound}
 		\norm{\bx - \gamma\bH^{\T}\bH\bx}_2 \leq \norm{\bI-\gamma\bH^{\T}\bH}_2\norm{\bx}_2,
 	\end{align}
 	where the spectral norm of $\bI -\gamma\HTH$ is given by
 	\begin{align}\label{eq:spectral_norm}
 		\norm{\bI-\gamma\bH^{\T}\bH}_2 = \max_{i}\{|1-\gamma\lambda_i|\}.
 	\end{align}
 	On the other hand, choosing $\gamma=\Frac{2}{(\lambda_{\max}+\lambda_{\min})}$ yields 
 	\begin{align}
 	&\frac{2\lambda_{\min}}{\lambda_{\max} + \lambda_{\min}} \leq \gamma\lambda_i \leq \frac{2\lambda_{\max}}{\lambda_{\max} + \lambda_{\min}},\quad \forall i \nonumber\\
 	\Leftrightarrow\quad& | 1-\gamma\lambda_i| \leq \frac{\lambda_{\max}-\lambda_{\min}}{\lambda_{\max} + \lambda_{\min}},\quad \forall i.\label{eq:absolute}
 	\end{align}
 	By combining~\eqref{eq:gradient_bound}, \eqref{eq:spectral_norm}, and~\eqref{eq:absolute}, 
 	\begin{align}\label{eq:gradient_Lipschitz}
 		\norm{\bx - \gamma\bH^{\T}\bH\bx}_2 \leq \frac{\lambda_{\max}-\lambda_{\min}}{\lambda_{\max} + \lambda_{\min}}\norm{\bx}_2, \quad\forall\bx.
 	\end{align}
 	Combining~\eqref{eq:gradient_Lipschitz}  with the Lipschitz continuity of $P_{\cS}$ gives
 	\begin{align}
 	&\norm{\G_{\gamma}(\bx) - \G_{\gamma}(\bz)}_2 \leq L \norm{(\bx - \bz) - \gamma\bH^{\T}\bH(\bx-\bz)}_2 \nonumber\\
 			 &\quad\leq L\,\frac{\lambda_{\max}-\lambda_{\min}}{\lambda_{\max} + \lambda_{\min}}\norm{\bx-\bz}_2,\quad \forall \bx,\forall\bz.\label{eq:contractive}
 	\end{align}
 	Since $L<(\lambda_{\max}+\lambda_{\min})/(\lambda_{\max}-\lambda_{\min})$, \eqref{eq:contractive} implies that $\G_{\gamma}$ is a contractive mapping. By the Banach-Picard fixed point theorem~\cite[Thm. 1.48]{BauschkeC:2011}, $\{\bx_k\}$ defined by $\bx_{k+1}=\G_{\gamma}(\bx_{k})$ converges to the unique fixed point $\bx^{*}$ of $\G_{\gamma}$,  for every  initialization $\bx_0$. Finally, since $P_{\cS}$ satisfies~\eqref{eq:local}, by Proposition~\ref{thm:minimizer}, $\bx^*$ is also a local minimizer of~\eqref{prob}.

 \subsection{Proof of Theorem~\ref{thm:fixed_point2}} \label{proof:fixed_point2}
 Again, let $\{\lambda_i\}$ be the set of eigenvalues of $\HTH$. With $\gamma<2/\lambda_{\max}$, one readily verifies that,  $\forall\bx\in\dimx$, 
 \begin{align*}
 \norm{\bx - \gamma\bH^{\T}\bH\bx}_2 &\leq \max_i\left\{|1-\gamma\lambda_i|\right\}\cdot\norm{\bx}_2\leq \norm{\bx}_2.
 \end{align*}
 Combining this with the non-expansiveness of $P_{\cS}$ leads to
 \begin{align*}
 \norm{\G_{\gamma}(\bx) - \G_{\gamma}(\bz)}_2 &\leq \norm{(\bx - \bz) - \gamma\bH^{\T}\bH(\bx-\bz)}_2\\
 &\leq \norm{\bx-\bz}_2, \quad\forall \bx,\bz\in\dimx.
 \end{align*}
 Now that $\G_{\gamma}$ is a non-expansive operator with a nonempty fixed-point set, we invoke the Krasnosel'ski\u{\i}-Mann theorem~\cite[Thm. 5.14]{BauschkeC:2011} to deduce that the iteration~\eqref{eq:averaged} must converge to a fixed point of $\G_{\gamma}$ which is, by Proposition~\ref{thm:minimizer}, also a local minimizer of~\eqref{prob}. 
 


\end{document}